\documentclass[conference]{IEEEtran}
\usepackage{times}

\usepackage[numbers]{natbib}
\usepackage{multicol}
\usepackage[bookmarks=true]{hyperref}
\usepackage[dvipsnames]{xcolor}
\usepackage{xfrac}

\usepackage{amsmath}
\usepackage{amssymb}
\usepackage{graphicx}
\usepackage{breqn}
\usepackage{subcaption}
\usepackage[font={footnotesize}]{caption}
\usepackage{xspace}
\usepackage[dvipsnames]{xcolor}
\usepackage{wrapfig}
\usepackage{relsize}

\usepackage{algorithm}
\usepackage{algcompatible}

\algnewcommand\algorithmicto{\textbf{to}}
\algnewcommand\RETURN{\State \textbf{return} }

\usepackage{hyperref}
\usepackage[capitalise]{cleveref}

\usepackage{array} 
\usepackage{tabularx}
\usepackage{multirow, makecell}
\usepackage{multicol}
\usepackage{booktabs}
\usepackage{pifont} 

\usepackage[
  separate-uncertainty = true,
  multi-part-units = repeat
]{siunitx}

\usepackage[hang,flushmargin]{footmisc}
\usepackage{url}

\usepackage[export]{adjustbox}

\usepackage{soul}

\usepackage{tablefootnote}
\newcommand{\name}{RoboCasa}

\begin{document}

\title{\texttt{RoboCasa}: Large-Scale Simulation of\\Everyday Tasks for Generalist Robots}

\author{\authorblockN{
Soroush Nasiriany$^{1}$, Abhiram Maddukuri$^{1, *}$, Lance Zhang$^{1, *}$, Adeet Parikh$^{1}$, \\Aaron Lo$^{1}$, Abhishek Joshi$^{1}$, Ajay Mandlekar$^{2}$, Yuke Zhu$^{1,2}$
}

\authorblockA{
\\
$^{1}$The University of Texas at Austin, $^{2}$NVIDIA Research; $^*$Denotes equal contribution
}

\authorblockN{
\\
\textcolor{RoyalBlue}{\textbf{\href{https://robocasa.ai}{robocasa.ai}}}
}

}


%

\makeatletter
\let\@oldmaketitle\@maketitle
\renewcommand{\@maketitle}{\@oldmaketitle
  \begin{center}
  \captionsetup{type=figure}
  \includegraphics[width=\textwidth]{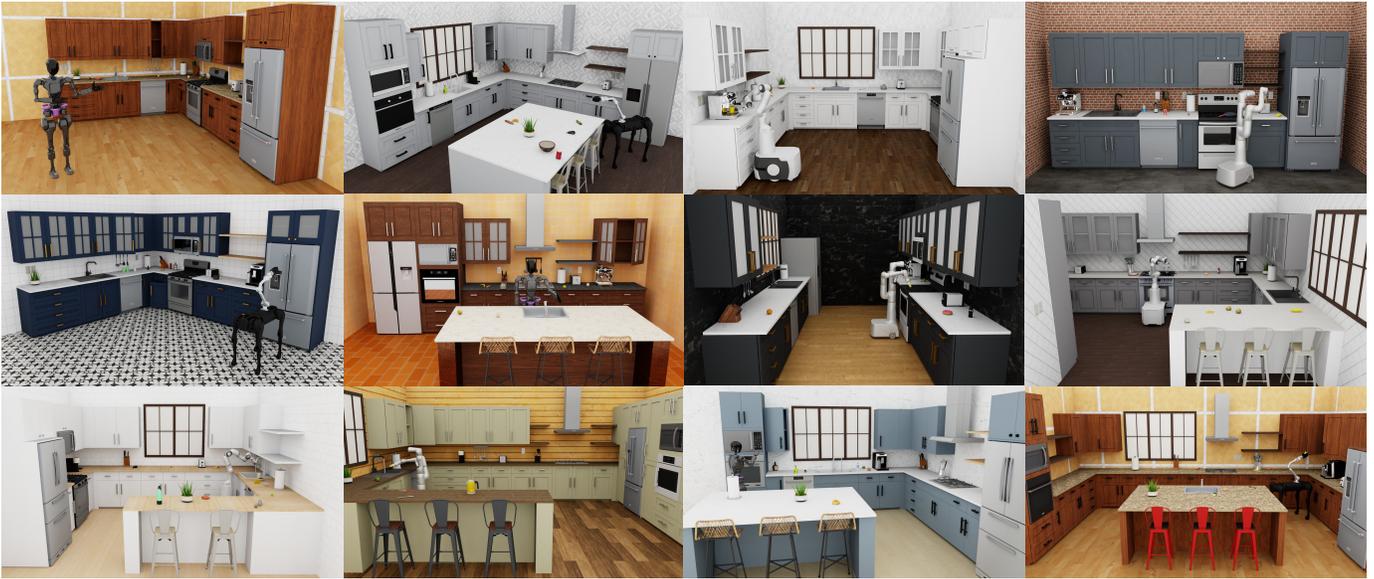}
    \captionof{figure}{\textbf{Overview of \name{}.} \name{} is a simulation framework for training generalist robot agents. Four pillars underlie \name{}: (1) Diverse assets, including 120 kitchen scenes and 2,500+ 3D objects, created with the aid of generative AI tools; (2) Cross-embodiment support for mobile manipulators and humanoid robots; (3) Diverse tasks created with the guidance of large language models; (4) Massive training datasets with over 100K trajectories.} 
    \label{fig:pull}
    \end{center}
    \vspace{-2mm}
}
\makeatother

\maketitle

\begin{abstract}
Recent advancements in Artificial Intelligence (AI) have largely been propelled by scaling.
In Robotics, scaling is hindered by the lack of access to massive robot datasets.
We advocate using realistic physical simulation as a means to scale environments, tasks, and datasets for robot learning methods.
We present \name{}, a large-scale simulation framework for training generalist robots in everyday environments.
\name{} features realistic and diverse scenes focusing on kitchen environments.
We provide thousands of 3D assets across over 150 object categories and dozens of interactable furniture and appliances.
We enrich the realism and diversity of our simulation with generative AI tools, such as object assets from text-to-3D models and environment textures from text-to-image models.
We design a set of 100 tasks for systematic evaluation, including composite tasks generated by the guidance of large language models.
To facilitate learning, we provide high-quality human demonstrations and integrate automated trajectory generation methods to substantially enlarge our datasets with minimal human burden.
Our experiments show a clear scaling trend in using synthetically generated robot data for large-scale imitation learning and show great promise in harnessing simulation data in real-world tasks. Videos and open-source code are available on the project website.
\end{abstract}

\IEEEpeerreviewmaketitle

\section{Introduction}
Recent breakthroughs in Artificial Intelligence have been driven by training giant neural network models on Internet-scale datasets. Unlike computer vision and natural language processing domains, where massive visual and text data are abundant from online sources, robotic data is relatively scarce. A key question in Robotics is how to acquire robotic training data that captures the vast diversity and complexity of the real world. Several prominent recent attempts have been made to create large, diverse datasets for training generalist robot models~\cite{rt12022arxiv, ebert2021bridge,open_x_embodiment_rt_x_2023,droid_2024}. While these datasets have advanced robots' generalization abilities in narrow domains, there remains a considerable gap between the capabilities achieved thus far and general-purpose robots that can be reliably deployed in the wild. It raises the question --- what is a viable path forward toward scaling in robot learning?

As collecting ever-larger datasets in the real world would require unrealistic amounts of capital and labor, many turn to simulation as a promising alternative to producing large quantities of synthetic data for model training. We expect simulation to play an integral role in scaling robot learning for the following reasons. 
First, once a feature-rich, high-fidelity simulator is created, we can generate large amounts of robot data at low cost. This is exemplified by recent automated data generation methods, such as MimicGen~\cite{mandlekar2023mimicgen} and Optimus~\cite{dalal2023imitating}, which exploit the privileged information of simulation to generate data with minimal human labor.
Second, the creation of realistic simulations has been facilitated by rapid advances in generative AI.
Today's generative AI tools are capable of generating images, synthesizing 3D assets, and writing source code~\cite{rombach2021highresolution,openai2024gpt4,geminiteam2024gemini}. These tools can be employed to create millions of scenes procedurally, import novel categories of objects, and program natural tasks and reward functions.
Finally, simulation democratizes and accelerates robot learning research, enabling rapid prototyping of new ideas and reproducible research.

To unleash the potential of simulation, it must satisfy three core criteria. First, the simulator must guarantee \textbf{realism} in physics, rendering, and underlying models to enable transfer to the real world.
Second, the simulator must satisfy \textbf{diversity} in the scenes, assets, and tasks it offers.
Generative AI will be crucial in enabling this diversity \textit{at scale}.
Finally, a simulator alone is not sufficient to train a highly capable generalist robot agent.
The simulation must be accompanied by \textbf{large robot datasets} that capture the diversity of scenes and behaviors that it has to offer.
Numerous prior attempts at creating simulations have partially satisfied some of these criteria, yet none have satisfied all.

We present \texttt{\name{}}, a large-scale simulation framework centered around home environments for training generalist robots.
\name{} builds upon RoboSuite~\cite{robosuite2020}, a modular, fast, and easy-to-use framework based in MuJoCo. \name{} inherits these features and goes far beyond by offering a large array of scenes, objects, and hardware platforms suited for building a general-purpose home robot. 
In this initial release, we focus our efforts on kitchen scenes.
To capture realistic and diverse scenes, we consult numerous architecture and home design magazines and compile several kitchen layouts and styles reflecting the diversity of kitchens in homes around the world.
We model these kitchens according to standard size and spatial specifications and fit them with a large repository of interactable furniture and appliances spanning cabinets, stoves, microwaves, coffee machines, and more.
Furthermore, we curate a repository of over 2,500 objects across over 150 categories, the majority of which are generated by text-to-3D tools.
\name{} has cross-embodiment support for mobile manipulators of diverse forms, such as single-arm mobile platforms, humanoid robots, and quadruped robots with arms. 

These assets allow us to simulate a wide range of behaviors in kitchen scenes. This release includes 100 tasks for systematic evaluation. The first 25 are atomic tasks that feature foundational robot skills, such as picking and placing, opening and closing doors, and twisting knobs. They serve as the basic building blocks to scaffold complex long-horizon tasks.
The other 75 are composite tasks involving a sequence of robot skills. We design these composite tasks to capture naturalistic kitchen activities by soliciting suggestions from large language models (LLMs).
Our key intuition is that LLMs are trained on human-centered Internet content, effectively capturing the ecological statistics of human behaviors.
We obtain a list of activities from LLMs, such as washing dishes, frying, and restocking cabinets.
Using these activities to ground our task design, we prompt the LLM to suggest concrete tasks for each activity. We look to expand the list of tasks in future releases.

We complement our tasks with high-quality human demonstrations across all 100 tasks. To augment our datasets, we extend MimicGen~\cite{mandlekar2023mimicgen} to generate 100K additional trajectories for our atomic tasks.
We train policies with behavioral cloning on human demonstrations and automatically generated data. We find that generated data significantly improves generalization, hinting at a promising path for scaling in robotics.
Furthermore, we show in a real-world kitchen environment that co-training with our simulation data significantly increases task success in real-robot deployment.
We summarize our contributions as follows:
\begin{itemize}
    \item We develop the \name{} simulation framework featuring diverse, realistic kitchen scenes, thousands of high-quality object assets, and cross-embodiment mobile manipulators. We employ generative AI tools to create environment textures and 3D objects.
    \item We introduce a set of 100 tasks for systematic evaluation, including 25 atomic tasks representing foundational sensorimotor skills and 75 composite tasks generated with the guidance of large language models.
    \item We provide a large multi-task dataset for model training, including large-scale synthetically generated trajectories. We show a clear scaling trend when using generated data and show the utility of simulation data in real-world tasks.
\end{itemize}

\section{Related Work}

\setlength{\tabcolsep}{0pt}
\definecolor{darkergreen}{RGB}{19,168,33}
\newcommand{\greencheck}{{\color{darkergreen}\ding{51}}}
\newcommand{\redcross}{{\color{red}\ding{55}}}

\begin{table*}[t!]
\centering
\resizebox{1.0\linewidth}{!}{

\begin{tabular}{l|c|cccccccccc}
\toprule
\textbf{Feature} & 
\begin{tabular}[c]{@{}c@{}}\rotatebox{30}{\textbf{RoboCasa}}\end{tabular} &
\begin{tabular}[c]{@{}c@{}}\rotatebox{30}{\textbf{AI2-THOR}}\end{tabular} & 
\begin{tabular}[c]{@{}c@{}}\rotatebox{30}{\textbf{Habitat 2.0}}\end{tabular} & 
\begin{tabular}[c]{@{}c@{}}\rotatebox{30}{\textbf{iGibson 2.0}}\end{tabular} &
\begin{tabular}[c]{@{}c@{}}\rotatebox{30}{\textbf{RLBench}}\end{tabular} & 
\begin{tabular}[c]{@{}c@{}}\rotatebox{30}{\textbf{Behavior-1K}}\end{tabular} & 
\begin{tabular}[c]{@{}c@{}}\rotatebox{30}{\textbf{robomimic}}\end{tabular} &
\begin{tabular}[c]{@{}c@{}}\rotatebox{30}{\textbf{ManiSkill 2}}\end{tabular} & 
\begin{tabular}[c]{@{}c@{}}\rotatebox{30}{\textbf{OPTIMUS}}\end{tabular} & 
\begin{tabular}[c]{@{}c@{}}\rotatebox{30}{\textbf{LIBERO}}\end{tabular} & 
\begin{tabular}[c]{@{}c@{}}\rotatebox{30}{\textbf{MimicGen}}\end{tabular} \\ 
\midrule

Mobile Manipulation & \greencheck & \greencheck & \greencheck & \greencheck & \redcross & \greencheck & \redcross & \greencheck & \redcross & \redcross & \greencheck \\
Room-Scale Scenes & \greencheck & \greencheck & \greencheck & \greencheck & \redcross & \greencheck & \redcross & \redcross & \redcross & \redcross & \redcross \\
Realistic Object Physics & \greencheck & \redcross & \redcross & \greencheck & \greencheck & \greencheck & \greencheck & \greencheck & \greencheck & \greencheck & \greencheck \\
AI-generated Tasks & \greencheck & \redcross & \redcross & \redcross & \redcross & \redcross & \redcross & \redcross & \redcross & \redcross & \redcross \\
AI-generated Assets & \greencheck & \redcross & \redcross & \redcross & \redcross & \redcross & \redcross & \redcross & \redcross & \redcross & \redcross \\
Photorealism & \greencheck & \greencheck & \greencheck & \redcross & \redcross & \greencheck & \redcross & \greencheck & \greencheck & \redcross & \redcross \\
Cross-Embodiment & \greencheck & \greencheck & \redcross & \greencheck & \redcross & \greencheck & \redcross & \redcross & \greencheck & \redcross & \greencheck \\
Num Tasks & 100 & - & 3 & 6 & 100 & 1000 & 8 & 20 & 10 & 130 & 12 \\
Num Scenes & 120 & - & 1 & 15 & 1 & 50 & 3 & - & 4 & 20 & 1 \\
Num Object Categories & 153 & - & 46 & - & 28 & 1265 & - & - & - & x & - \\
Num Objects & 2509 & 3578 & 169 & 1217 & 28 & 5215 & 15 & 2144 & 72 & x & 40 \\
Human Data & \greencheck & \redcross & \redcross & \greencheck & \redcross & \redcross & \greencheck & \redcross & \redcross & \greencheck & \greencheck \\
Machine-Generated Data & \greencheck & \redcross & \redcross & \redcross & \greencheck & \redcross & \greencheck & \greencheck & \greencheck & \redcross & \greencheck \\
Num Trajectories & 100K+ & - & - & - & - & 0 & 6K & 30K & 245K & 5K & 50K \\

\bottomrule
\end{tabular}

}
\vspace{+3pt}
\caption{\footnotesize{\textbf{Comparison to Popular Simulation Frameworks used in the Robot Learning Literature.}}}
\label{table:sim_comparison}
\end{table*}

\textbf{Simulation Frameworks for Robotics.} Many simulation frameworks have been built for robotics --- we provide a thorough comparison between \name{} and popular frameworks in Table~\ref{table:sim_comparison}. Some are limited to tabletop manipulation~\cite{robomimic2021, james2020rlbench, dalal2023imitating, liu2023libero}, but \name{}, along with others~\cite{kolve2017ai2, szot2021habitat, li2021igibson, li2023behavior, gu2023maniskill2, mandlekar2023mimicgen}, support mobile manipulation. \name{} also supports room-scale scenes, unlike other frameworks that include mobile manipulation in smaller portions of a room~\cite{gu2023maniskill2, mandlekar2023mimicgen}. \name{} runs realistic physics for all interactions, including object grasping and placement, unlike some other mobile manipulation frameworks such as AI2-THOR~\cite{kolve2017ai2} and Habitat 2.0~\cite{szot2021habitat}. \name{} is one of the few frameworks to feature photorealistic rendering~\cite{kolve2017ai2, szot2021habitat, li2023behavior, gu2023maniskill2, dalal2023imitating} and multiple robot embodiments~\cite{kolve2017ai2, li2021igibson, li2023behavior, dalal2023imitating, mandlekar2023mimicgen}. \name{} also has a large collection of tasks, room-scale scenes, and objects --- only a small number of other works~\cite{li2023behavior,szot2023large} offer these at scale.
Meanwhile, recent work in incorporating generative AI tools have explored AI-generated tasks~\cite{wang2023gen} and scene configurations~\cite{wang2023robogen}.
But critically, \name{} is the only one to support a large array of tasks, room-scale scenes, and objects while incorporating AI-generated tasks and assets, ensuring potentially limitless diversity in scenes and tasks.
Furthermore, unlike many other simulation platforms (including Behavior-1K~\cite{li2023behavior}) we provide large-scale datasets of task demonstrations through a combination of human teleoperation and the MimicGen system~\cite{mandlekar2023mimicgen} (more discussion below) and provide a thorough analysis of agents trained via imitation learning across our large collection of tasks.  The convergence of diverse scenes, tasks, and assets alongside the extensive dataset provided by \name{} will fulfill a crucial requirement not addressed by any other simulation frameworks in the robot learning community.

\textbf{Datasets and Benchmarks for Robotics.} 
Recently, several large-scale data collection efforts have been made for robotics. 
One approach is self-supervised learning, where trial-and-error is used to collect data for tasks like grasping and pushing~\cite{levine2016learning, pinto2016supersizing, kalashnikov2018qt, kalashnikov2021mt, yu2016more, dasari2019robonet}. This can take significant time to generate high-quality data due to the trial-and-error process.
A popular alternative is to collect robot demonstrations via human teleoperation, where a human controls a robot to guide it through different tasks~\cite{zhang2017deep, mandlekar2018roboturk, mandlekar2019scaling, mandlekar2020hitl,  ebert2021bridge, jang2022bcz}. Several recent efforts have scaled this paradigm up by using teams of human operators over extended periods of time~\cite{ebert2021bridge, ahn2022can, jang2022bcz, rt12022arxiv}. However, most of these efforts focus on collected real-world datasets. By contrast, we focus on collecting large-scale datasets in simulation, where results are easier to reproduce and thorough evaluations are possible due to lower human burdens.

Another approach is to leverage algorithmic trajectory generators in simulation~\cite{james2020rlbench, zeng2020transporter, jiang2023vima, gu2023maniskill2, dalal2023imitating}, but these efforts often make use of privileged information and hand-designed heuristics, and can consequently be difficult to apply to arbitrary tasks without significant human effort. Some recent efforts have used Large Language Models to generate datasets in simulation~\cite{wang2023robogen, ha2023scaling}, but it can still involve carefully engineered pipelines. 
To combine the quality and wide applicability of human teleoperation with the scale of using pre-programmed demonstrators in simulation, we collect a set of human demonstrations in simulation and then leverage MimicGen~\cite{mandlekar2023mimicgen}. This recently proposed data generation system synthesizes additional demonstrations using a set of human demonstrations to generate a much larger dataset.

\textbf{Learning from Large Offline Datasets.}
Behavioral Cloning~\cite{pomerleau1989alvinn} is a popular method for learning policies offline from a set of demonstrations. It trains a policy to imitate the actions in the dataset. It has been used extensively in prior works~\cite{zhang2017deep, mandlekar2020learning, ebert2021bridge, rt12022arxiv, jang2022bcz, dalal2023imitating, jiang2023vima}. 
Offline Reinforcement Learning~\cite{levine2020offline} is an alternative method that tries to prefer certain dataset actions over others using a reward function. It has also been used to learn from large offline robot manipulation datasets~\cite{kalashnikov2021mt, chebotar2023q, gurtler2023benchmarking, kumar2022pre, kumar2021workflow}.
In this work, we use Behavioral Cloning using a Transformer-based~\cite{vaswani2017transformer} visuomotor policy similar to other works~\cite{dalal2023imitating} to train agents on large offline datasets. We also consider other popular policy architectures, namely diffusion models~\cite{ho2020denoising, chi2023diffusion}.

\begin{figure*}[t!]
    \centering
    \includegraphics[width=1.0\linewidth]{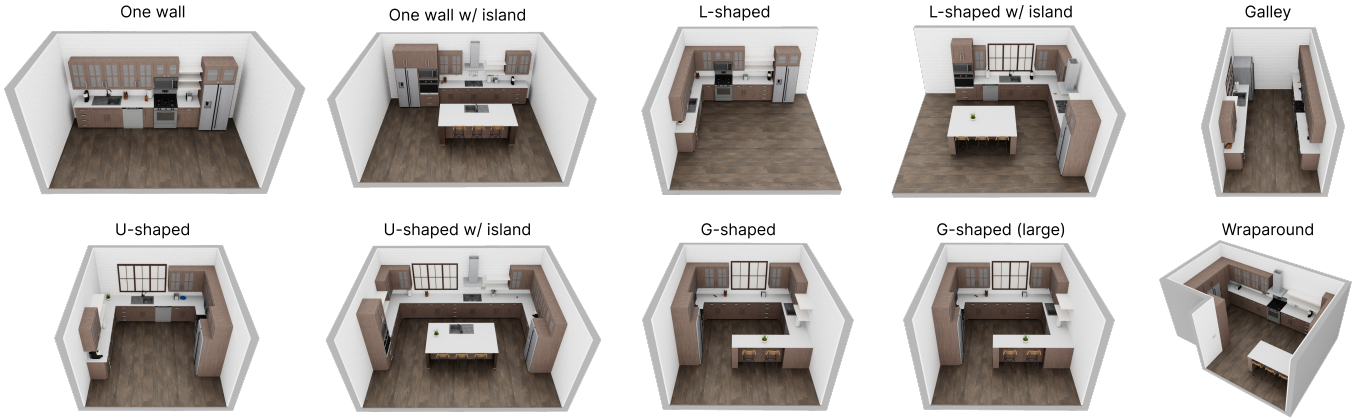}
    \caption{\textbf{Kitchen Floor Plans.} We consult home planning and architecture magazines and compile a list of common kitchen floor plans. Our floor plans take on a variety of shapes and sizes, from basic designs (\textit{e.g.}, one wall) to high-end ones (\textit{e.g.}, U-shaped w/ island).}
    \label{fig:layout-blueprints}
\end{figure*}
\section{RoboCasa Simulation}
We outline the core simulation components of \name{}. We highlight our efforts to create diverse and realistic kitchen scenes, furniture and appliances, and objects.

\subsection{Core Simulation Platform}
We adopt RoboSuite~\cite{robosuite2020} as the core simulation platform on which we develop \name{}.
We chose RoboSuite because of its focus on physical realism, high speed, and modular design, which allows us to scale to large-scale scenes.
We directly inherit several core components of RoboSuite, including the environment model formats and robot controllers.
Crucially, in order to support room-scale environments, we extend RoboSuite to accommodate mobile manipulators, including robots mounted on wheeled bases, humanoid robots, and quadrupeds with arms.
We obtain and adapt these models from various robotics repositories~\cite{robosuite2020,menagerie2022github,haviland2022holistic}.
We also support high-quality rendering with NVIDIA Omniverse, allowing us to capture photorealistic images (see~\Cref{fig:pull}).

\subsection{Kitchen Scenes}
In this initial release, we focus on household tasks centered around kitchen activities.
We created a large array of kitchen scenes with fully interactive cabinets, drawers, and appliances.
We consulted online home design and architecture magazines to compile a diverse list of kitchen floor plans.
We model 10 floor plans (\Cref{fig:layout-blueprints}) ranging from basic designs found in apartments to more elaborate designs found in high-end homes.
Each kitchen can be configured to take on a custom architectural style.
After consulting architecture magazines, we compile the popular kitchen styles, including Industrial, Scandinavian, Coastal, Modern, Traditional, Mediterranean, Rustic, and more.
Each style features a unique combination of design elements, including textures, appliance choices, and cabinet panels and handles.
For example, Scandinavian kitchens employ light, low-contrast textures and simple, sleek cabinet panels and appliances.
In contrast, Mediterranean kitchens use ornate appliances, glass panel cabinets, and colorful textures.
In total, we have modeled 12 kitchen styles, and we showcase these styles across different floor plans in~\Cref{fig:pull}.
Each floor plan can be configured to take on any style, resulting in 120 kitchen scenes.
Each scene can be customized further by replacing textures from a large selection of high-quality AI-generated textures.
We have 100 textures for walls, 100 for the floor, 100 for counters, and 100 for cabinet panels.
We use the popular text-to-image tool MidJourney to generate these images.
We use these textures as a form of domain randomization to significantly increase the visual diversity of our training datasets.

\begin{figure}[!t]
    \centering
    \vspace{0.5em}
    \includegraphics[width=1.0\linewidth]{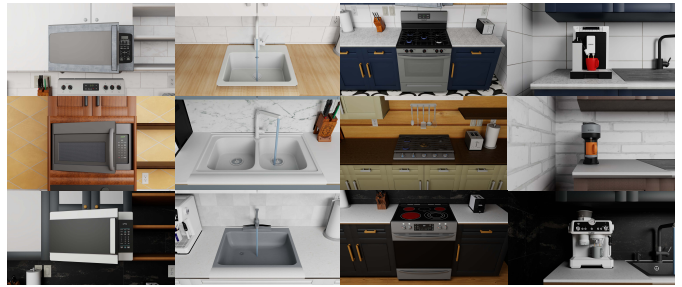}
    \caption{\textbf{Examples of Interactable Appliances.} Our simulation framework comes with dozens of appliances. Several types of appliances are articulated. For example, we can open and close doors on microwaves and twist knobs on stoves. Some appliances can undergo state changes, \textit{e.g.}, when we turn the knob on the stove, the corresponding burner turns on.}
    \label{fig:appliances}
\end{figure}

\subsection{Assets}
We create a large repository of intractable 3D assets to accommodate diverse kitchen activities.
Our repository includes cabinets, drawers, and various kitchen appliances.
We source these assets from 3D model repositories online and convert them to the MuJoCo MJCF model format. Our postprocessing operation involves segmenting appliances into articulated entities, for example, segmenting the door of a microwave and the knobs on a stove. It allows us to represent rich interactions, such as closing a microwave door or turning on a stove.
Furthermore, these appliances undergo state changes, \textit{e.g.}, when we turn a stove knob on, the corresponding burner turns on to simulate heat.
See \Cref{fig:appliances} for an illustration of our appliances.

\begin{figure}[!t]
    \centering
    \vspace{0.5em}
    \includegraphics[width=1.0\linewidth]{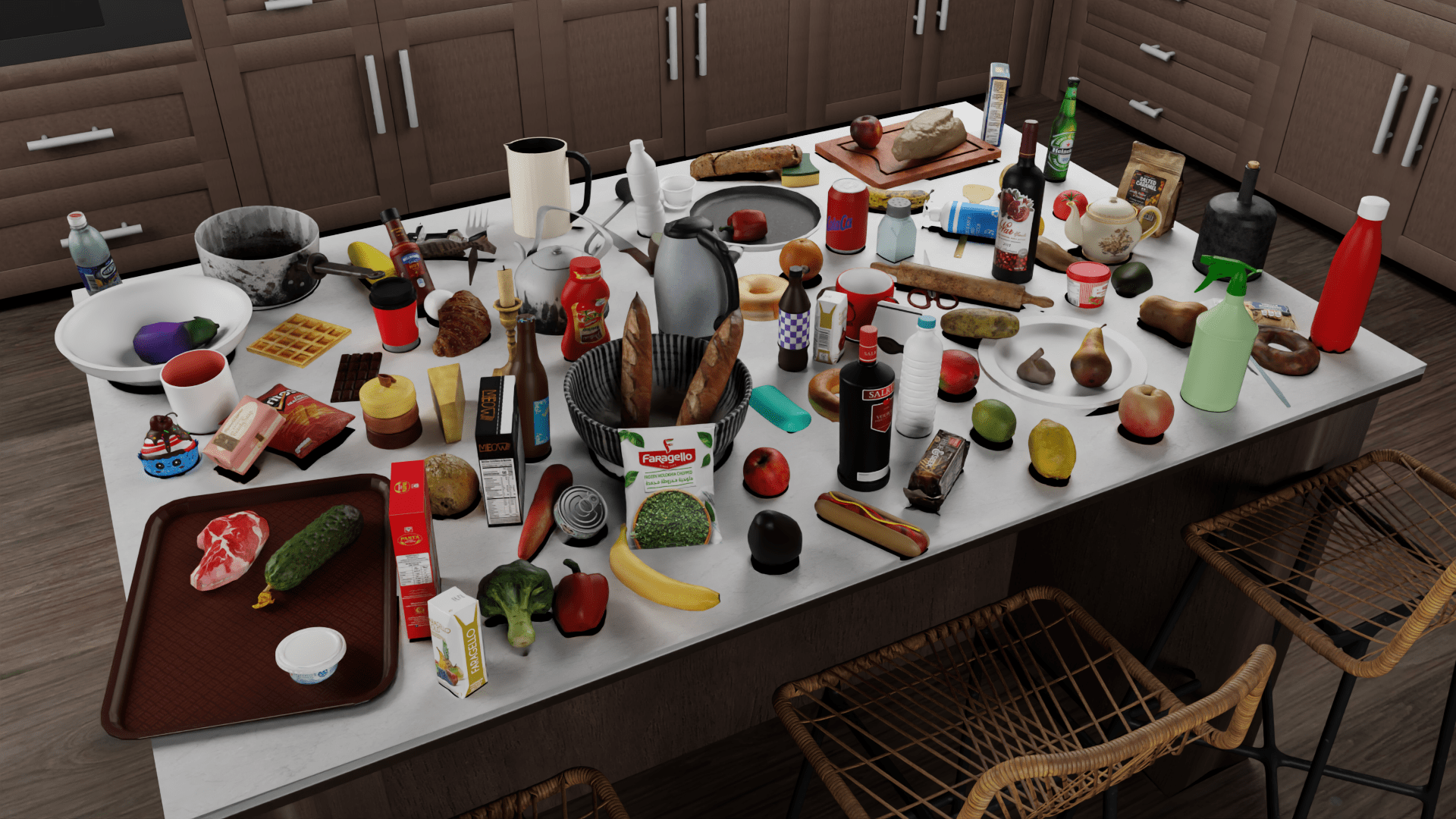}
    \caption{\textbf{Diverse High-Quality 3D Objects.} \name{} offers 2,509 high-quality 3D objects across 153 diverse categories spanning vegetables, poultry, drinks, and more. Here we illustrate a small subset of these objects.}
    \label{fig:objects}
    \vspace{-1em}
\end{figure}

In addition to appliances, we create a rich library of objects commonly found in kitchens, spanning fruits and vegetables, dairy, poultry, drinks, receptacles, tools, and more.
We gather object assets from two sources, the Objaverse~\cite{objaverse} dataset and Luma.ai, an online text-to-3D service.
We mine a large set of candidate objects and filter out defective or low-quality ones.
At the end of this process, we collect 2,509 high-quality assets spanning 153 unique object categories.
The majority of these assets (1,592) are sourced from Luma.ai.
See \Cref{fig:objects} for an illustration of our objects.

\begin{figure*}[!t]
    \centering
    \includegraphics[width=1.0\linewidth]{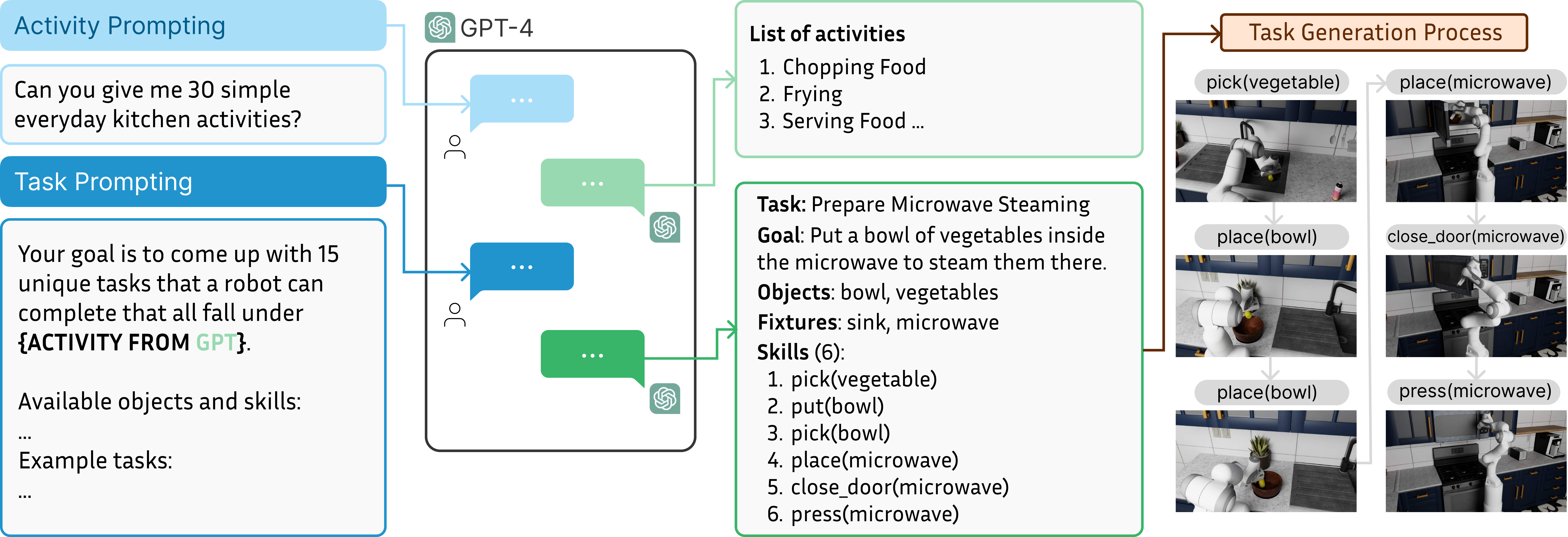}
    \caption{\textbf{Creating Diverse Tasks with Large Language Models.} We employ LLMs to generate diverse tasks. First, we prompt GPT-4 to give diverse high-level kitchen activities. Subsequently, for each activity, we prompt GPT-4 (or Gemini 1.5) to suggest a diverse set of representative tasks. We illustrate one such task ``Preparing the Microwave For Steaming" for the activity label ``Steaming Vegetables".}
    \label{fig:gpt-tasks}
\end{figure*}

\section{\name{} Activity Dataset}
\label{sec:dataset}
Our simulator supports a wide array of possible kitchen activities, and we represent these activities with a comprehensive suite of 100 tasks. This section outlines these tasks and our large multi-task dataset accompanying them.

\subsection{Atomic Tasks: Building Blocks of Behavior}
\label{subsec:atomic_tasks}
For a robot to perform complex tasks, it must master the foundational skills needed to solve these tasks.
We focus on a set of eight sensorimotor skills that form the basis for the majority of household activities: 1) Pick and place, 2) Opening and closing doors, 3) Opening and closing drawers, 4) Twisting knobs, 5) Turning levers, 6) Pressing buttons, 7) Insertion, and 8) Navigation.
These skills do not constitute an exhaustive list, and including additional skills centered around behaviors such as deformable manipulation is left for future work.
To effectively learn the skills, we propose a set of 25 tasks that each involve one of these eight skills.
We will refer to these as \textit{atomic tasks}.
The full breakdown of these tasks is outlined in the appendix (\Cref{fig:atomic-tasks-table}).

\subsection{Creating Composite Tasks with Large Language Models}
\label{subsec:composite_tasks}
Our composite tasks involve sequencing skills to solve semantically meaningful activities such as cooking and cleaning.
Our goal in creating these tasks is to capture diverse tasks that reflect the ecological statistics of real-world household activities.
We use the guidance of large language models (LLMs) to define our tasks.
This approach offers several key benefits.
First, LLMs encapsulate diverse sources of human knowledge and can thus effectively communicate diverse ideas grounded in the real world.
In addition, these LLMs can be used at scale to define thousands of unique tasks, significantly reducing the human labor involved in task definition.
We generate tasks across two steps (see~\Cref{fig:gpt-tasks}).
First, we prompt ChatGPT (GPT-4~\cite{openai2024gpt4}) to list common high-level kitchen activities.
We compile a list of 20 activities: \textit{brewing coffee or tea}, \textit{washing dishes}, \textit{restocking kitchen supplies}, \textit{chopping food}, \textit{making toast}, \textit{defrosting food}, \textit{boiling water}, \textit{meat preparation}, \textit{setting the table}, \textit{clearing the table}, \textit{sanitizing}, \textit{snack preparation}, \textit{tidying cabinets and drawers}, \textit{washing fruits and vegetables}, \textit{frying}, \textit{reheating food}, \textit{mixing and blending}, \textit{baking}, \textit{serving food}, and \textit{steaming vegetables}.
We then prompt GPT-4 and Gemini 1.5~\cite{geminiteam2024gemini} to propose representative tasks for each activity label.
The LLMs occasionally exhibit logical flaws, so we filter or modify some of their outputs.
We compile 75 task \textit{blueprints} in total from the LLM and proceed to code implementations for them.
Except for a select number of composite tasks designed to work in certain environments, all tasks are simulatable in any of our kitchen scenes.
We describe in detail our prompts and tasks in the appendix.

\subsection{\name{} Datasets}
\label{subsec:datasets}

We have outlined a comprehensive set of 100 tasks consisting of 25 atomic tasks (Sec.~\ref{subsec:atomic_tasks}) and 75 composite tasks created with LLMs (Sec.~\ref{subsec:composite_tasks}). This section explains how we collect our large-scale demonstration dataset across these tasks. We first use human teleoperation to collect a base set of demonstrations and then use automated trajectory generation methods to expand this to a much larger set of demonstrations. 

\textbf{Collecting a base set of demonstrations through human teleoperation.} A team of four human operators collect 50 high-quality demonstrations for each atomic task using a 3D SpaceMouse~\cite{zhu2018reinforcement, robosuite2020}. Each task demonstration is collected in a random kitchen scene (random kitchen floor plan, random kitchen style, and random AI-generated textures). This results in large and diverse simulation datasets through human teleoperation (1,250 demonstrations). However, our experiments show that even this scale of human data is insufficient to solve most of our tasks. This is likely due to the significant scope and diversity of the tasks and scenes. Consequently, we opt to use data generation tools to expand our data quantity.

\textbf{Leveraging automated trajectory generation methods to synthesize demonstrations.} To further scale the dataset's size with minimal human effort, we employ MimicGen~\cite{mandlekar2023mimicgen}, a recently developed trajectory generation method. MimicGen can automatically synthesize rich datasets from a seed set of human demonstrations by adapting them to new settings. The core generation mechanism first decomposes each human demonstration into a sequence of object-centric manipulation segments. Then, for a novel scene, it transforms each object-centric segment according to the current pose of the relevant object, stitches the segments together, and has the robot follow the new trajectory to collect a new task demonstration. 

MimicGen requires some basic assumptions on simulation. We outline how they are easily satisfied in \name{}: MimicGen assumes that tasks consist of a known sequence of object-centric subtasks --- this sequence must be specified for each new task. Fortunately, the atomic tasks (Sec.~\ref{subsec:atomic_tasks}) consist of eight core skills. All tasks that correspond to a skill have the same or similar sequences of object-centric subtasks, with the main differences coming from the identity of the reference object. For example, for pick-and-place tasks, the first stage is a pick subtask with one reference object, and the second stage is a place subtask with a second reference object. Consequently, specifying subtask sequences takes minimal human effort. In addition, each human demonstration provided to MimicGen must also be annotated with segments corresponding to each object-centric subtask. This can be done with automated metrics that detect the end of each subtask --- these functions only need to be implemented once for each of the eight core skills and can be re-used across the entire set of demonstrations.

MimicGen data generation attempts are not always successful. In practice, it employs a rejection sampling scheme only to keep generation attempts that lead to task success. Leveraging RoboCasa simulation, we parallelize MimicGen data generation across multiple simulation processes to speed up the data generation process.
\section{Experiments}

We aim to explore the following research questions in our experiments:
\begin{enumerate}
    \item How effective are machine-generated trajectories from MimicGen in learning multi-task policies, in comparison to human demonstrations?
    \item How will the generalization performance of the imitation learning policy scale with increasing training dataset sizes?
    \item Can large-scale simulation datasets facilitate knowledge transfer to downstream tasks within simulation and facilitate policy learning for real-world tasks?
\end{enumerate}

\begin{figure*}[!t]
    \centering
    \includegraphics[width=1.0\linewidth]{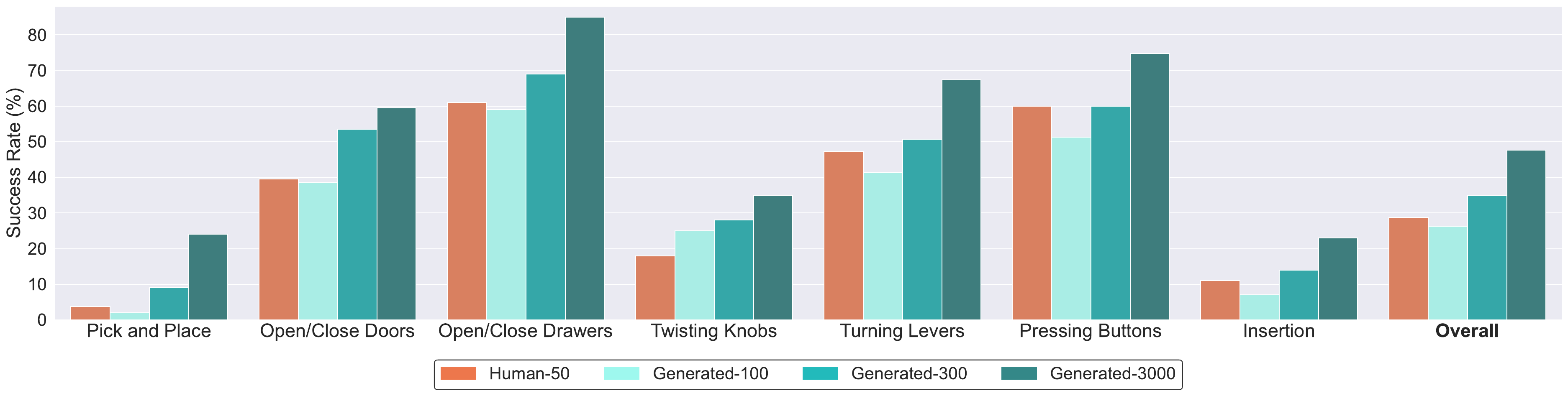}
    \caption{\textbf{Comparison between human demonstrations and machine-generated datasets.} We present learning results across 24 atomic tasks spanning diverse robot skills. We compare training on four different multi-task datasets, including a human dataset with 50 demonstrations per task, a machine generated dataset with 3000 demonstrations per task, and smaller variants with 300 or 100 demonstrations per task. We group task results according to their corresponding sensorimotor skills (see ~\Cref{fig:main-results-table} for a full breakdown of results by task). We see a clear scaling trend: increasing the size of the generated dataset can yield consistently higher overall success rates, eventually significantly outperforming performance on human datasets.}
    \label{fig:main-results}
\end{figure*}

\subsection{Imitation Learning for Atomic Tasks}
First, we perform a systematic study with the atomic tasks. One question we are interested in is how the performance of imitation learning policies compare when trained on human data versus machine-generated data and how the scale of this data plays a role in performance. We compare the following four multi-task data settings:
\vspace{1mm}
\begin{enumerate}
    \item \textbf{Human-50}: A dataset of 1250 human demonstrations spanning all 25 atomic tasks, each with 50 human demonstrations.\vspace{1mm}
    \item \textbf{Generated-3000}: A dataset of 72,000 demonstrations synthesized by MimicGen\footnote{These experiments feature Objaverse objects. We release an additional 28K trajectories featuring AI-generated objects, forming the full 100K trajectory dataset.} across 24 atomic tasks\footnote{We exclude the kitchen navigation task, as MimicGen is not currently able to generate mobile manipulation trajectories. We leave this to future work.} We take the 50 human demonstrations as input for each task and use them to generate 3,000 trajectories autonomously. \vspace{1mm}
    \item \textbf{Generated-300}: A random $\sfrac{1}{10}$ subset  of our full generated dataset, where we generate 300 demos per task. This results in a total of 7,200 trajectories.\vspace{1mm}
    \item \textbf{Generated-100}: A random $\sfrac{1}{30}$ subset of our full generated dataset, where we generate 100 demos per task. This results in a total of 2,400 trajectories.
\end{enumerate}
\vspace{1mm}
Images in the training datasets are rendered with AI-generated textures. We evaluate our policies in kitchen scenes with human-curated textures.
In these datasets, our focus is specifically on a Franka Panda robot with an Omron mobile base, resembling the Omni-Frankie robot~\cite{haviland2022holistic}. 
We train a visuomotor policy with behavioral cloning on each of these four multi-task datasets.
We specifically use the publicly available BC-Transformer implementation in RoboMimic~\cite{robomimic2021}. Refer to \Cref{app:policy-learning} for additional details.

After training, we perform a comprehensive evaluation of the model. For each task, we evaluate the model performance across 50 trials across five fixed evaluation scenes, each with a distinct floor plan and style. In order to test generalization capabilities, we only evaluate the policy only on unseen object instances.
Additionally, two of the five scenes encompass unseen styles that were never encountered in the training data.
We report results in~\Cref{fig:main-results} where we group results together with tasks belonging to the same skill.
The overall performance on human data is $28.8\%$ success rate, and with the fully generated dataset, we observe a significant improvement at $47.6\%$ success rate.
Furthermore, we observe a \textit{scaling trend} from using machine-generated data: as we increase the quantity of generated data, the model performance increases steadily.
This offers a promising outlook: data generation tools enable us to learn significantly more performant agents at a relatively low cost.
We observe several other notable findings in the results.
Some skills are significantly easier to learn (\textit{e.g.}, opening and closing doors and drawers), while others are quite challenging (\textit{e.g.}, pick and place).
We hypothesize a number of factors explaining these findings. First, tasks exhibiting high diversity are significantly more challenging to learn.
One example is the pick and place tasks involving dozens of different object categories with a wide range of affordances.
In comparison, opening and closing doors involves six different instances of doors and is thus significantly easier to learn.
Another factor is dexterity, as we see that tasks involving high levels of dexterity, such as insertion, are challenging to learn. 

\subsection{Imitation Learning for Composite Tasks}
Next we study learning on our composite tasks.
These tasks are more challenging as they require multiple skills, which increases the horizon of the task and introduces new subtleties.
Due to the increased difficulty of these tasks and the challenges of multi-task learning, we opt to learn a single-task policy for each task.
For each task, we collected 50 human demonstrations and compared the following settings:
\vspace{1mm}
\begin{itemize}
    \item \textbf{Scratch}: learning a policy from scratch on these 50 demonstrations;\vspace{1mm}
    \item \textbf{Fine-tuning}: taking our pre-trained policy learned on atomic tasks with the full MimicGen generated dataset and fine-tuning on these 50 demonstrations.
\end{itemize}
\vspace{1mm}
We independently train models for the following five tasks:
\vspace{1mm}
\begin{itemize}
    \item \texttt{ArrangeVegetables}: This task belongs to the ``chopping food" activity. The robot must place two vegetables from the sink onto the cutting board on the counter;\vspace{1mm}
    \item \texttt{MicrowaveThawing}: This task belongs to the ``defrosting food" activity. The robot must place a frozen food item from the counter inside the microwave and turn on the microwave;\vspace{1mm}
    \item \texttt{RestockPantry}: This task belongs to the ``restocking kitchen supplies" activity. The robot must pick and place multiple cans from the counter to the cabinet. There are a number of cans already in the cabinet, among other objects. The robot must locate the existing cans in the cabinet (either on the right or left side) and place the new cans right next to them;\vspace{1mm}
    \item \texttt{PreSoakPan}: This task belongs to the ``washing dishes" activity. The robot must pick and place a pan and a sponge into the sink and turn on the water faucet to prepare the pan for washing;\vspace{1mm}
    \item \texttt{PrepareCoffee}: This task belongs to the ``brewing coffee of tea" activity. The robot must take a mug out of the cabinet, place it under the coffee machine, and press the coffee machine button to serve it into the mug.
\end{itemize}
\vspace{1mm}

\begin{figure}[!t]
\centering
\small
\begin{tabular}{l | c| c } 
 \toprule
  & $\quad$Scratch$\quad$ & $\quad$Fine-tuning$\quad$ \\
 \midrule
\texttt{ArrangeVegetables} & $2.0\%$  & $12.0\%$ \\
\texttt{MicrowaveThawing} & $0\%$  & $2.0\%$ \\
\texttt{RestockPantry} & $0\%$  & $6.0\%$ \\
\texttt{PreSoakPan} & $0\%$  & $4.0\%$ \\
\texttt{PrepareCoffee} & $0\%$  & $0\%$ \\
\bottomrule
\end{tabular}
\caption{\textbf{Learning Results on Composite Tasks.} We learn single-task policies for five representative composite tasks. We compare learning these tasks from scratch with 50 human demonstrations versus fine-tuning a policy trained on machine-generated atomic task data. The fine-tuning method performs better but still struggles to learn robust behaviors.}
\label{fig:composite-task-results}
\end{figure}

See~\Cref{fig:composite-task-results} for results. Learning on these composite tasks is very challenging, with the Scratch baseline failing to achieve any non-zero success rate on 4/5 tasks.
The fine-tuning method achieves non-zero success rates on 4/5 tasks.
Some common failure modes include difficulty with fine-grained manipulation and difficulty effectively transitioning to the next stage of the task.
However, we generally observe that the fine-tuned models perform better qualitatively, with more robust picking and placing strategies in particular.
We attribute this to the large pretraining dataset of atomic behaviors.
Our benchmark leaves room for significant improvement on these tasks.
The choice of policy architecture, learning algorithm, and fine-tuning strategy may play a critical role in performance, and these factors warrant investigation in future work.

\subsection{Transfer to Real World Environments}
We show how large-scale data generated in simulation can aid in learning tasks in RoboCasa and other domains, including in the real world. We conduct experiments in a real-world kitchen environment with a Franka Emika Panda robot running on the DROID hardware infrastructure~\cite{droid_2024}.
While both the real world and simulated Franka robot are controlled via workspace end effector control, our simulated robot uses Operational Space Control while the DROID-based real robot does not.
In addition, our robot controller runs at 20 Hz frequency while the real robot controller runs at 15 Hz.
In addition to controller differences, there are differences in camera calibration, lighting conditions, and the placement of the robot base with respect to the scene.

Our experiments include three tasks in the real kitchen environment: 1) pick and place an object from the counter to the sink, 2) pick and place an object from the sink to the counter, and 3) pick and place an object from the counter to the cabinet.
These tasks resemble single-stage tasks in RoboCasa. For each task, we collected 50 demonstrations, each over five distinct object categories.
We train a policy for each task, and we compare the following two settings:
\vspace{1mm}
\begin{itemize}
    \item \textbf{Real only}: training on the real-world demos for the target task only;\vspace{1mm}
    \item \textbf{Real\,+\,Sim}: co-training on real-world demos for the target task and all of our simulation MimicGen demonstrations over all single-stage tasks.
\end{itemize}
\vspace{1mm}
In \Cref{fig:real-robot}, we report policy success rates (mean and standard deviation, in percentage) averaged over 3 seeds.
For each seed, we evaluate the model over five seen object categories and 3 unseen object categories (unseen with respect to the real-world demonstrations).
On seen objects, we see that co-training with simulated data yields a 24.4\% average success rate, compared to 13.6\% with using real data only, a relative improvement of 
79\%.
While performance suffers on unseen objects, we still see a significant improvement in incorporating simulation data.
We attribute this to the rich diversity and visual and physical realism of our simulator.

\begin{figure}[t!]
\centering
\includegraphics[width=1.0\linewidth]{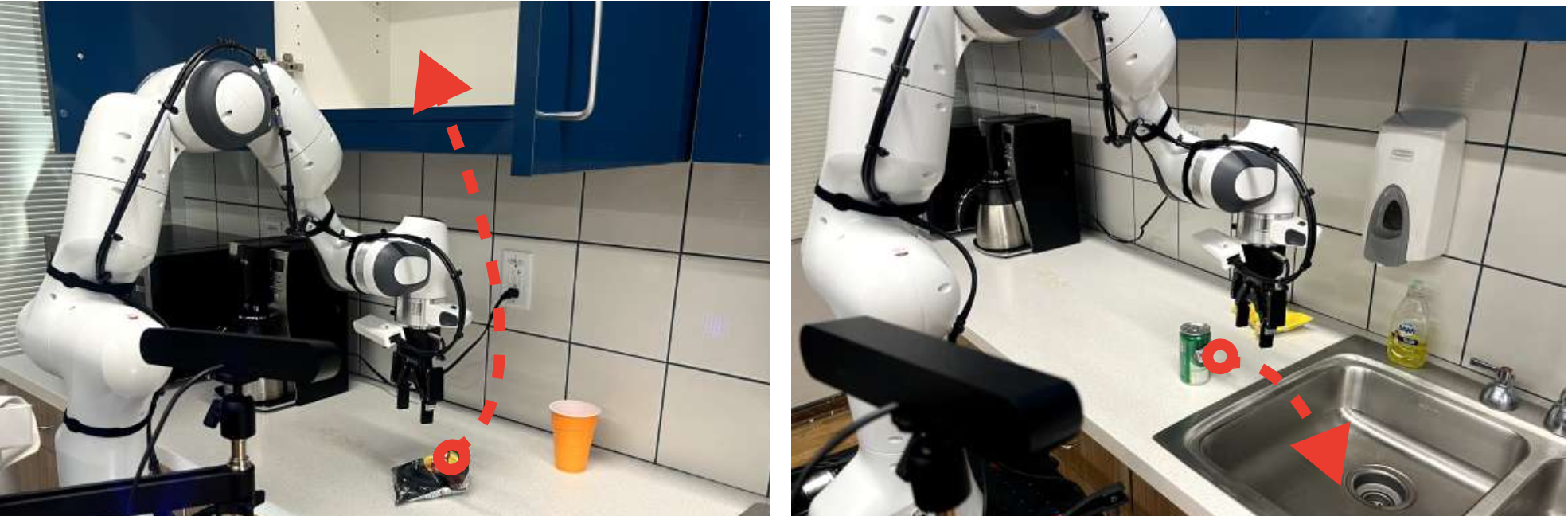} 
\caption{\textbf{Real-World Experiment Setup.} We conduct experiments in a real-world kitchen environment with a Franka Emika Panda arm on a wheeled mobile platform. The two pictures illustrate two of the three evaluation tasks: (left) pick and place an object from the counter to the cabinet, and (right) pick and place an object from the counter to the sink.}
\label{fig:real-robot-setup}
\label{fig:real-robot-setup}
\end{figure}

\begin{figure}[!t]
\centering
\small
\centering
\begin{tabular}{l|l|c|c}
\toprule
\textbf{Setting} & \textbf{\;Task} & \;\textbf{Real only\;} & \;\textbf{Real\,+\,Sim (Ours)\;} \\
\midrule
\multirow{2}{*}{Seen Obj} & \;Counter to sink & 12.7 $\pm$ 2.5 & \textbf{22.0 $\pm$ 2.8} \\
& \;Sink to counter  & 20.0 $\pm$ 5.9 & \textbf{29.3 $\pm$ 4.1} \\
& \;Counter to cabinet\;  & 8.0 $\pm$ 1.6 & \textbf{22.0 $\pm$ 5.8} \\ \cmidrule{2-4}
& \;Task average  & 13.6 & \textbf{24.4} \\
\midrule

\multirow{2}{*}{Unseen Obj\;} & \;Counter to sink & 3.3 $\pm$ 4.7 & \textbf{8.9 $\pm$ 7.9}\\
& \;Sink to counter & 1.1 $\pm$ 1.6 & \textbf{7.8 $\pm$ 4.2} \\
& \;Counter to cabinet\; & 3.3 $\pm$ 4.7 & \textbf{11.1 $\pm$ 11.0} \\ \cmidrule{2-4}
& \;Task average  & 2.6 & \textbf{9.3} \\

\bottomrule
\end{tabular}
\caption{\textbf{Real Robot Evaluations.} In a real-world kitchen domain with only a handful of demonstrations, we explore co-training policies with our simulation data. Compared to training policies exclusively on in-domain real-world demonstrations, co-training substantially improves policy performance.}
\label{fig:real-robot}
\end{figure}

\vspace{3mm}
\section{Conclusion}
We have presented \name{}, a large-scale simulation framework for training generalist robots in everyday environments.
\name{} features 120 realistic scenes, dozens of appliances, 2,500+ high-quality 3D objects spanning 150+ categories, 100 diverse tasks, and a large multi-task dataset of 100K+ trajectories.
Generative AI tools play a central role in our simulation, with object assets from text-to-3D models, environment textures from text-to-image models, and LLMs to generate kitchen activities and corresponding tasks.
In our experiments, we show that synthetically generated data in simulation can be useful in scaling robot policy learning.

We now pinpoint limitations and discuss exciting avenues for future future.
First, our experiments show that fine-tuning on composite tasks yields relatively low performance, leaving room for improvement.
In the future, we will investigate more powerful policy architectures and learning algorithms and improve the quality of our machine-generated datasets. 
While the generated trajectories are technically considered successful, many exhibited undesirable effects, such as jerky motions and collisions.
Many of these behaviors can be automatically detected by checking simulation states and trajectories exhibiting such behaviors can be discarded.
Next, while we showed how to use LLMs to create tasks, this process still required human guidance to write the implementations for these tasks.
One interesting research direction is to use LLMs to propose thousands of new scenes and tasks and write code to implement these scenes and tasks with minimal human guidance.
We anticipate that this will be possible as LLMs become more performant.
In addition, we would like to extend the scope beyond kitchen environments and tasks in future releases.
Next, our dataset consists of critical coarse manipulation behaviors but does not encapsulate highly dexterous skills, deformable manipulation tasks, or tasks that require bimanual manipulation.
Finally, we are interested in exploring training using a combination of data from our simulation, other simulators, and diverse sources such as internet videos and real robot datasets.
We envision a future in which these diverse forms of data complement each other to create a powerful foundation model for robotics.

\section*{Acknowledgments}
We would like to thank Yifeng Zhu for significant contributions in the cross-embodiment efforts. We also thank Yuqi Xie for rendering support and Roberto Martín-Martín for discussions on mobile manipulation support. Finally, we thank all UT Austin Robot Perception and Learning Lab members for their invaluable feedback on RoboCasa. This work has been partially supported by the National Science Foundation (FRR-2145283, EFRI-2318065) and the Office of Naval Research (N00014-22-1-2204).


\bibliographystyle{plainnat}
\bibliography{references}





\newpage
\newpage
\newpage
\newpage
\clearpage
\section{Simulator}
We benchmark the speed of our simulator on the \texttt{PickPlaceCounterToCab} task, running for 10 episodes, with each episode spawned in a random scene.
We use native MuJoCo rendering on an NVIDIA RTX A5000 GPU which adds overhead to the simulation speed.
The simulation physics runs on CPUs and we specifically used an AMD EPYC 7543 32-Core Processor.
The average time to reset the scene is $9.50$ seconds and the average speed of stepping through the simulator (calling \texttt{env.step}) is $25.2$ fps.
Without rendering, the average reset time is $9.46$ seconds and the simulator speed is $31.9$ fps.
For reference, each timestep in our simulator corresponds to $0.04$ seconds in the world or $25$ fps. Our observed simulation speed with the rendering of $25.2$ fps almost exactly matches this speed, meaning our simulator runs roughly at real-time speed.
\section{Tasks and Datasets}

\subsection{Atomic Tasks}
\label{app:atomic-tasks}
We design 25 atomic tasks representing eight foundational robot skills: (1) Pick and place, (2) Opening and closing doors, (3) Opening and closing drawers, (4) Twisting knobs, (5) Turning levers, (6) Pressing buttons, (7) Insertion, and (8) Navigation.
We outline details for all tasks in \Cref{fig:atomic-tasks-table}.
Each task can come in the form of multiple task variants.
These task variants can disambiguate the task goal with language.
For pick-and-place tasks, there is a task variant for each object category being manipulated to disambiguate objects in clutter.
For turning on and off the stove, there is a task variant for each burner being turned on or off, to disambiguate the relevant burner. 
For navigation, there is a task variant for each appliance the robot is navigating to, to disambiguate the target location.

\subsection{Composite Tasks}
\label{app:composite-tasks}
We obtain activity labels by asking ChatGPT the following simple prompt:
``Can you give me 30 simple everyday high-level kitchen activities? Each activity should be unique.''
We obtain a set of candidate responses and manually select 20 activities.
We use a more elaborate prompt to obtain task suggestions.
We list the robot skills, relevant object categories and fixtures, specify constraints of the simulation (eg. small objects that cannot be grasped, limited support for deformable manipulation tasks), and highlight example task blueprints as a form of few-shot prompting.
We list all 20 kitchen activities and representative tasks across all activities in \Cref{fig:composite-tasks-table}.

In selecting composite tasks, we filter out LLM task suggestions that have logical flaws. For a list of examples, see below:

\noindent \newline Using invalid object (no blender exists): \\
\begin{small}
\texttt{
Task: Set Up Blending Station \\
Goal: Place dairy ingredients next to the \colorbox{pink}{blender} for making creamy fillings or batter. 
Objects: cheese, milk  \\
Fixtures: counter, blender \\
Skills (3): Pick_up(dairy), Place(counter), Push(dairy, blender)  \\
Reasoning: Preparing dairy products near a blender is common when making mixtures for baking.
}
\end{small}

\noindent \newline Improper use of skills: \\
\begin{small}
\texttt{
Task: Wine Selection for Cooking \\
Goal: Retrieve a bottle of wine from the cabinet for use in a recipe \\
Objects: drink (wine) \\
Fixtures: cabinet
Skills (5): Open(cabinet), Pick_up(drink), Place(counter), Close(cabinet), \colorbox{pink}{Press(button_on_coffee_machine)}(simulating uncorking) \\
Reasoning: Wine is occasionally used in baking recipes and needs to be opened and ready to use.
}
\end{small}

\noindent \newline Picking up objects that should not be grasped (utensils):
\begin{small}
\texttt{
Task: Retrieve Baking Utensils \\
Goal: Gather all utensils needed for baking (spoons, ladle) and place them on the countertop. \\
Objects: utensils \\
Fixtures: drawer, counter \\
Skills (4): Open(drawer), \colorbox{pink}{Pick_up(utensils)}, Place(counter), Close(drawer) \\
Reasoning: Assembling the correct utensils is essential before starting to bake.
}
\end{small}

\subsection{Datasets}
In our atomic task experiments, we use human datasets and machine-generated datasets over 25 tasks.
We render images for these tasks using randomly sampled AI-generated textures, which replace the native textures for each scene.
Due to the high volume of these datasets and time constraints, we opt to use the light-weight MuJoCo renderer to render images for these datasets.
For the public release we will provide users the option to render all datasets with the Omniverse renderer.
\section{Policy Learning Implementation}
\label{app:policy-learning}
Our BC-Transformer policy takes as input the history of the past 10 observations in addition to the language goal for the text and outputs the next ten actions for the robot to execute.
The agent replans after executing the first action.
We modified the policy to support language conditioning by encoding language goals using a CLIP sentence encoder.
For each observation in the input, the policy encodes proprioceptive information (end-effector pose and mobile base pose) and images from three cameras: an eye-in-hand camera, a left workspace camera, and a right workspace camera.
It encodes each of these images with a dedicated ResNet-18 encoder stack and fuses the visual representation using FiLM layers.
The encoded observations are passed to a 6-layer Transformer with $\sim 20M$ trainable parameters.
We train the model for 500k gradient steps at a learning rate of $1e-4$ with a learning rate warmup.

We also experiment with diffusion policy~\cite{chi2023diffusion}.
We implement the diffusion policy in RoboMimic for a fair comparison to our existing BC-Transformer method. The diffusion policy uses the same observation encoder (ResNet, FiLM conditioning) as the BC-Transformer. We use all recommended hyperparameters from the official implementation: 2 timesteps for observation history, a prediction horizon of 16 steps, and an action horizon of 8 steps.
We use DDIM~\cite{song2022denoising} with 100 train timesteps and 10 inference timesteps, as recommended by the Diffusion Policy authors.
We found the Diffusion Policy to underperform the BC-Transformer implementation significantly. On the single-stage PickPlaceCounterToSink task, BC-Transformer achieves a 56\% success rate while Diffusion Policy only achieves 12\%. One possible explanation for this is that our BC-Transformer implementation uses a longer history length of 10 observations, while diffusion policy uses a history length of 2 observations (this is the default choice used by Chi et al. which we also opt to use). Incorporating a longer observation history may be critical for our tasks.

\begin{figure*}
\centering
\small
\setlength{\tabcolsep}{2pt}
\begin{tabular}{l | m{2.7cm} | m{8cm}}
 \toprule
  Task & Skill Family & $\quad$Description$\quad$\\
 \midrule
\texttt{PickPlaceCounterToCabinet} & pick and place  & Pick an object from the counter and place it inside the cabinet. The cabinet is already open.\\[4mm]
\texttt{PickPlaceCabinetToCounter} & pick and place  & Pick an object from the cabinet and place it on the counter. The cabinet is already open.\\[4mm]
\texttt{PickPlaceCounterToSink} & pick and place  & Pick an object from the counter and place it in the sink.\\[4mm]
\texttt{PickPlaceSinkToCounter} & pick and place  & Pick an object from the sink and place it on the counter area next to the sink.\\[4mm]
\texttt{PickPlaceCounterToMicrowave} & pick and place  & Pick an object from the counter and place it inside the microwave. The microwave door is already open.\\[4mm]
\texttt{PickPlaceMicrowaveToCounter} & pick and place  & Pick an object from inside the microwave and place it on the counter. The microwave door is already open.\\[4mm]
\texttt{PickPlaceCounterToStove} & pick and place  & Pick an object from the counter and place it in a pan or pot on the stove.\\[4mm]
\texttt{PickPlaceStoveToCounter} & pick and place  & Pick an object from the stove (via a pot or pan) and place it on (the plate on) the counter.\\[4mm]
\texttt{OpenSingleDoor} & opening and closing doors  & Open a microwave door or a cabinet with a single door.\\[4mm]
\texttt{CloseSingleDoor} & opening and closing doors  & Close a microwave door or a cabinet with a single door.\\[4mm]
\texttt{OpenDoubleDoor} & opening and closing doors  & Open a cabinet with two opposite-facing doors.\\[4mm]
\texttt{CloseDoubleDoor} & opening and closing doors  & Close a cabinet with two opposite-facing doors.\\[4mm]
\texttt{OpenDrawer} & opening and closing drawers  & Open a drawer.\\[4mm]
\texttt{CloseDrawer} & opening and closing drawers  & Close a drawer.\\[4mm]
\texttt{TurnOnStove} & twisting knobs  & Turn on a specified stove burner by twisting the respective stove knob.\\[4mm]
\texttt{TurnOffStove} & twisting knobs  & Turn off a specified stove burner by twisting the respective stove knob.\\[4mm]
\texttt{TurnOnSinkFacuet} & turning levers  & Turn on the sink faucet to begin the flow of water.\\[4mm]
\texttt{TurnOffSinkFaucet} & turning levers  & Turn off the sink faucet to begin the flow of water.\\[4mm]
\texttt{TurnSinkSpout} & turning levers  & Turn the sink spout.\\[4mm]
\texttt{CoffeePressButton} & pressing buttons  & Press the button on the coffee machine to pour coffee in to the mug.\\[4mm]
\texttt{TurnOnMicrowave} & pressing buttons  & Turn on the microwave by pressing the start button.\\[4mm]
\texttt{TurnOffMicrowave} & pressing buttons  & Turn off the microwave by pressing the stop button.\\[4mm]
\texttt{CoffeeSetupMug} & insertion & Pick the mug from the counter and insert it onto the coffee machine mug holder area.\\[4mm]
\texttt{CoffeeServeMug} & insertion & Remove the mug from the coffee machine mug holder and place it on the counter.\\[4mm]
\texttt{NavigateKitchen} & navigation & Navigate to a specified appliance in the kitchen.\\[4mm]
\bottomrule
\end{tabular}
\caption{\textbf{Atomic Tasks.}}
\label{fig:atomic-tasks-table}
\end{figure*}
\begin{figure*}
\centering
\small
\setlength{\tabcolsep}{2pt}
\begin{tabular}{l | m{3cm} | m{8cm}} 
 \toprule
  Task & Activity & Description \\
 \midrule
\texttt{PrepareCoffee} & Brewing coffee or tea  & Place a mug under the coffee machine nozzle and press the start button to make coffee. \\[3mm]
\texttt{DryDishes} & Washing dishes & Set bowls and cups on the counter to dry, after they have been washed. \\[3mm]
\texttt{RestockPantry} & Restocking kitchen supplies & Group all canned foods together on a shelf in the cabinet, right next to existing canned foods. \\[3mm]
\texttt{ArrangeVegetables} & Chopping food & Place vegetables
from the sink onto the cutting board on the counter, to prepare for chopping them. \\[3mm]
\texttt{CheesyBread} & Making toast & Pick up the wedge of cheese and place it on the slice of bread. \\[3mm]
\texttt{MicrowaveThawing} & Defrosting food & Pick frozen food and place in microwave, then turn on microwave to thaw the food. \\[3mm]
\texttt{FillKettle} & Boiling water &  Place an empty kettle from the cabinet and place in the sink to be filled with water. \\[3mm]
\texttt{PrepMarinatingMeat} & Meat preparation & Place steak and condiments on cutting board for marination. \\[3mm]
\texttt{DateNite} & Setting up the table & Pick up candles and a bottle of wine and place them in the dining area for date night. \\[3mm]
\texttt{BowlAndCup} & Clearing the table & An empty bowl and an empty cup are left on the island. Stack the cup inside the bowl and move the stacked items to the counter. \\[3mm]
\texttt{PrepForSanitizing} & Sanitizing & Organize cleaning supplies on the counter for easy access. \\[3mm]
\texttt{MakeFruitBowl} & Snack preparation & Gather various fruits into a bowl for serving. \\[3mm]
\texttt{PantryMishap} & Tidying cabinets and drawers &  Sort the canned and packaged foods into a drawer while placing the vegetables on a nearby counter. \\[3mm]
\texttt{DrainVeggies} & Washing fruits and vegetables & Dump the veggies in the pot into the sink to drain. Turn off the faucet, and place the pot back on the counter. \\[3mm]
\texttt{SetupFrying} & Frying & Retrieve a pan and transport it to the stove, then turn on the stove. \\[3mm]
\texttt{HeatMug} & Reheating food & Place a mug in the microwave and then turn on the microwave. \\[3mm]
\texttt{ColorfulSalsa} & Mixing and blending & Create salsa by gathering specific vegetables from a mixed pile on the counter and placing them onto the cutting board. \\[3mm]
\texttt{OrganizeBakingIngredients} & Baking & Obtain dairy ingredients and place them together for baking.  \\[3mm]
\texttt{PlaceFoodInBowls} & Serving food & Arrange bowls on the counter and place an assortment of vegetable in each bowl. \\[3mm]
\texttt{SteamInMicrowave} & Steaming vegetables & Put a bowl of vegetables inside the microwave to steam them there. \\[3mm]
\bottomrule
\end{tabular}
\caption{\textbf{Representative Composite Tasks Across 20 Kitchen Activites.}}
\label{fig:composite-tasks-table}
\end{figure*}

\begin{figure*}
\centering
\setlength{\tabcolsep}{2pt}

\setlength{\tabcolsep}{0.5em} 
{\renewcommand{\arraystretch}{1.5}
\begin{tabular}{|l|l|l|l|l|}
\hline
                               & \textbf{Human-50} & \textbf{Generated-100} & \textbf{Generated-300} & \textbf{Generated-3000} \\ \hline
\textbf{PickPlaceCabToCounter}       & 0.02              & 0.04                   & 0.10                    & \textbf{0.18}           \\ \hline
\textbf{PickPlaceCounterToCab}       & 0.06              & 0.08                   & 0.16                   & \textbf{0.28}           \\ \hline
\textbf{PickPlaceCounterToMicrowave} & 0.02              & 0                      & 0                      & \textbf{0.18}           \\ \hline
\textbf{PickPlaceCounterToSink}      & 0.02              & 0.02                   & 0.16                   & \textbf{0.44}           \\ \hline
\textbf{PickPlaceCounterToStove}     & 0.02              & 0                      & 0                      & \textbf{0.06}           \\ \hline
\textbf{PickPlaceMicrowaveToCounter} & 0.02              & 0                      & \textbf{0.12}          & 0.08                    \\ \hline
\textbf{PickPlaceSinkToCounter}      & 0.08              & 0.02                   & 0.14                   & \textbf{0.42}           \\ \hline
\textbf{PickPlaceStoveToCounter}     & 0.06              & 0                      & 0.04                   & \textbf{0.28}           \\ \hline
\textbf{OpenSingleDoor}        & 0.46              & 0.42                   & 0.44                   & \textbf{0.50}            \\ \hline
\textbf{OpenDoubleDoor}        & 0.28              & 0.12                   & 0.22                   & \textbf{0.48}           \\ \hline
\textbf{CloseDoubleDoor}       & 0.28              & 0.18                   & \textbf{0.62}          & 0.46                    \\ \hline
\textbf{CloseSingleDoor}       & 0.56              & 0.82                   & 0.86                   & \textbf{0.94}           \\ \hline
\textbf{OpenDrawer}            & 0.42              & 0.26                   & 0.40                    & \textbf{0.74}           \\ \hline
\textbf{CloseDrawer}           & 0.8               & 0.92                   & \textbf{0.98}          & 0.96                    \\ \hline
\textbf{TurnOnStove}           & 0.32              & 0.42                   & 0.44                   & \textbf{0.46}           \\ \hline
\textbf{TurnOffStove}          & 0.04              & 0.08                   & 0.12                   & \textbf{0.24}           \\ \hline
\textbf{TurnOnSinkFaucet}      & 0.38              & 0.26                   & \textbf{0.48}          & 0.34                    \\ \hline
\textbf{TurnOffSinkFaucet}     & 0.50               & 0.48                   & 0.46                   & \textbf{0.72}           \\ \hline
\textbf{TurnSinkSpout}         & 0.54              & 0.50                    & 0.58                   & \textbf{0.96}           \\ \hline
\textbf{CoffeePressButton}     & 0.48              & 0.48                   & 0.42                   & \textbf{0.74}           \\ \hline
\textbf{TurnOnMicrowave}       & 0.62              & 0.36                   & 0.76                   & \textbf{0.90}            \\ \hline
\textbf{TurnOffMicrowave}      & \textbf{0.70}      & \textbf{0.70}           & 0.62                   & 0.60                     \\ \hline
\textbf{CoffeeServeMug}        & 0.22              & 0.12                   & 0.24                   & \textbf{0.34}           \\ \hline
\textbf{CoffeeSetupMug}        & 0                 & 0.02                   & 0.04                   & 0.12                    \\ \hline
\textbf{Average}               & 0.288             & 0.263                  & 0.350                  & \textbf{0.476}          \\ \hline
\end{tabular}
}

\caption{\textbf{Multi-task Learning on Atomic Tasks: Full Results}}
\label{fig:main-results-table}
\end{figure*}

\end{document}